\documentclass[letterpaper, 9 pt, conference]{ieeeconf}  % Comment this line out if you need a4paper

\IEEEoverridecommandlockouts                              % This command is only needed if 
\usepackage{booktabs}
\usepackage{graphicx}
\usepackage{multirow}
\usepackage{amsmath,amssymb,amsfonts}
\usepackage{amsthm}
\usepackage{cite}
\newtheorem{theorem}{Theorem}

\newtheorem{assumption}{Assumption}
\usepackage[utf8]{inputenc}
\let\labelindent\relax
\usepackage[inline]{enumitem}
\usepackage{xcolor}
\usepackage{tikz}
\usetikzlibrary{positioning,arrows.meta}
\usepackage{algorithm}
\usepackage{algpseudocode}
\usepackage{pgfplots}
\pgfplotsset{compat=1.17}

\title{\LARGE \bf
A Causal Information-Flow Framework for Unbiased Learning-to-Rank
}
\author{Haoming Gong$^{*}$, Qingyao Ai$^{\dagger}$, Zhihao Tao$^{*}$, and Yongfeng Zhang$^{*}$\\
$^{*}$Department of Computer Science, Rutgers University, New Brunswick, NJ, USA\\
$^{\dagger}$Department of Computer Science and Technology, Tsinghua University, Beijing, China}

\begin{document}

\maketitle
\thispagestyle{empty}
\pagestyle{empty}

%%%%%%%%%%%%%%%%%%%%%%%%%%%%%%%%%%%%%%%%%%%%%%%%%%%%%%%%%%%%%%%%%%%%%%%%%%%%%%%%
\begin{abstract}
In web search and recommendation systems, user clicks are widely used to train ranking models. However, click data is heavily biased, i.e., users tend to click higher-ranked items (position bias), choose only what was shown to them (selection bias), and trust top results more (trust bias). Without explicitly modeling these biases, the true relevance of ranked items cannot be correctly learned from clicks. Existing Unbiased Learning-to-Rank (ULTR) methods mainly correct position bias and rely on propensity estimation, but they cannot measure remaining bias, provide risk guarantees, or jointly handle multiple bias sources. To overcome these challenges, this paper introduces a novel causal learning-based ranking framework that extends ULTR by combining Structural Causal Models (SCMs) with information-theoretic tools. SCMs specify how clicks are generated and help identify the true relevance signal from click data, while conditional mutual information, measures how much bias leaks into the learned relevance estimates. We use this leakage measure to define a rigorous notion of disentanglement and include it as a regularizer during model training to reduce bias. In addition, we incorporate a causal inference estimator, i.e., 
doubly robust estimator, to ensure more reliable risk estimation. Experiments on standard Learning-to-Rank benchmarks show that our method consistently reduces measured bias leakage and improves ranking performance, especially in realistic scenarios where multiple biases-such as position and trust bias-interact strongly.

\end{abstract}

%%%%%%%%%%%%%%%%%%%%%%%%%%%%%%%%%%%%%%%%%%%%%%%%%%%%%%%%%%%%%%%%%%%%%%%%%%%%%%%%
\section{Introduction}

The utilization of user feedback in training ranking models is constrained by inherent biases in user browsing behavior and document presentation. Specifically, users tend to exhibit position bias, where they have a higher likelihood of clicking on relatively higher-ranked documents \cite{Xuanhui01}. Additionally, selection bias \cite{wang2016learning}, trust bias \cite{agarwal2019addressing}, and outlier bias \cite{sarvi2023impact} further contribute to the potential bias in the training of these models. These biases mix non-relevance signals into the learned scores, making it challenging to recover true relevance from observed clicks alone. Because examination and trust depend on both the ranking and the content, the missingness of relevance labels induced by clicks is typically not missing at random. The empirical ranking risk derived from this type of noisy labels is biased and insufficient for sound supervised learning.

Understanding and correcting these biases is critical for
  contemporary ranking systems, such as web search and recommendation
  systems \cite{liu2009learning,joachims2002optimizing}. Without
  proper debiasing, models trained on biased clicks will
  perpetuate and potentially amplify the biases present in the
  training data, leading to suboptimal user experiences and poor
  ranking quality when deployed. This
  challenge has motivated extensive research into methods that can
   recover unbiased relevance estimates from biased user feedback
  \cite{joachims2017unbiased,ai2018unbiased}.

Classical click models provide one approach to understanding user behavior. These models parameterize user examination and decision behavior, for example via position-based, cascade, or dynamic Bayesian models that relate exposure, perceived relevance, and clicks \cite{craswell2008experimental,chapelle2009dynamic,chuklin2015click}. These models capture important behavioral regularities (e.g., reduced attention at lower ranks) and, when fitted, can be used to de-bias click-through rates or construct synthetic supervision signals. However, they are typically defined as observational generative models and, even with excellent fit to logged clicks, they do not by themselves yield unbiased or identifiable estimates of ranking risk for new deployment policies.

The Unbiased Learning-to-Rank (ULTR) framework instead takes a causal perspective on click data. It views position bias as a systematic distortion of clicks and attempts to correct it by estimating, for each query--document and position, how likely the document was to be exposed, then reweighting clicks by the inverse of this propensity \cite{joachims2017unbiased}. Under suitable assumptions, this inverse propensity score (IPS) weighting  makes the estimated ranking loss match the loss we would observe if exposure were unbiased. Dual Learning Algorithm (DLA) \cite{ai2018unbiased} strengthens this idea by jointly learning a ranker and a position-based propensity model, recognizing that accurate relevance estimation requires accurate propensity estimation and vice versa. Later work improves propensity estimation and stability under strong logging policies \cite{agarwal2019addressing,oostpolicyerhuis2020} and introduces conditional mutual information objectives that explicitly reduce dependence between position and learned scores \cite{wang2020information}.

Despite these advances, current ULTR methods face several important limitations that hinder their practical deployment and theoretical understanding. One key issue is the lack of quantitative measurement: propensity-based methods typically do not measure how much bias remains in the learned relevance signal after reweighting. Information-theoretic approaches \cite{wang2020information} minimize observational dependence (correlation in observed data) but do not ensure causal independence (independence under intervention) or provide bounds on ranking risk. This lack of quantitative measurement makes it difficult for practitioners to assess whether debiasing has actually succeeded.
% A related limitation is the absence of a clear connection between bias levels and ranking performance. 
Also, existing work does not provide guarantees that relate the amount of residual bias in the learned model to the degradation in ranking quality. In other words, if we measure some level of bias leakage in our model, we cannot say how much this will hurt the model's ranking performance in deployment. As a result, practitioners lack principled guidance on how to interpret a reported bias level in terms of ranking performance, or how to allocate limited debiasing resources across different bias mechanisms when multiple biases are active simultaneously. Furthermore, most existing methods focus on a single bias channel and fold all user-behavior effects into a scalar propensity, making it difficult to model and mitigate multiple bias mechanisms (e.g., position and trust) jointly.

In this work, we address these limitations by adopting a structural information-flow perspective on unbiased learning from clicks. Our starting observation is that unbiased learning is ultimately about disentangling true relevance from user-behavior bias. At the level of the data-generating process, relevance and bias become entangled through the pathways that connect context features, ranking scores, exposure, trust, and clicks. We therefore adopt a structural causal model (SCM) that makes these pathways explicit and distinguishes between the logging regime (the deployed system) and a target regime in which selected bias channels, such as exposure or trust, are set to idealized behaviors (for example, always-exposed). Unbiasedness can then be phrased as a requirement that the learned ranking scores be independent of these bias channels.

Rather than assuming that reweighting removes bias, we treat bias as structural information flow in the proposed framework: the information that actually travels from upstream bias sources (logging policy, presentation, etc.) into a small set of bias channels (exposure, decision/trust, presentation) and ultimately into the learned scores. Mutual information along these channels quantifies the remaining entanglement between relevance and bias and provides a natural notion of bias intensity. We also provide theoretical analyses, showing that the excess risk incurred by training under the logging regime can be controlled by the total information that flows along these channels, turning leakage into a measurable quantity that can be regularized. Because the same SCM can capture multiple channels simultaneously, this view extends seamlessly from pure position bias to settings with combined position and trust bias without extra complexity.
% Our technical approach formulates unbiased LTR as an SCM in which bias sources (logging policy, presentation) influence clicks through a small set of bias channels (exposure, post-exposure trust). 
Moreover, our training objective augments a standard LTR loss with mutual-information penalties that quantify leakage from bias sources into these channels. This yields quantitative risk bounds: the gap between logging-regime risk and target-regime risk is controlled by the square root of total channel leakage. Classical ULTR emerges as the single-channel special case; our framework extends naturally to multi-channel settings by penalizing each channel's leakage independently. For robustness to model misspecification, we employ doubly robust risk estimation.

Empirically, we validate our approach on standard ranking benchmarks, i.e., Yahoo! LETOR and MSLR-WEB30K, under both position-only and position+trust bias settings. Under pure position bias on Yahoo! LETOR, our method (SIF) achieves NDCG@10 = 0.756, outperforming the best baseline (Pair-debias: 0.751). On MSLR-WEB30K under position bias, SIF achieves NDCG@10 = 0.406, outperforming the second-best methods (IPS/DLA at 0.391) by 1.5 percentage points, demonstrating consistent effectiveness across datasets of different scale and feature spaces. Under multi-channel position+trust bias on Yahoo! LETOR, SIF improves NDCG@10 by 2.5 percentage points over DLA (0.740 vs 0.715), demonstrating the importance of multi-channel debiasing. Ablation studies confirm that our MI regularization effectively reduces measured leakage during training and improves exposure propensity estimation accuracy, with learned exposure probabilities achieving mean absolute error (MAE) of 0.237 to the oracle PBM versus 0.248 for DLA.

% In summary, this paper makes the following contributions:
% \begin{itemize}[leftmargin=*]
%   \item \textbf{Framework.}
%         A unified SCM + structural information-flow formulation of ULTR that treats
%         bias as information flow through explicit channels and operationalizes
%         unbiasedness as (approximate) interventional information-independence or
%         disentanglement between relevance and bias.
%   \item \textbf{Theory.}
%         Risk--divergence and divergence--leakage bounds that connect measurable
%         bias leakage to excess ranking risk, including an exposure-only specialization
%         and a principled extension to multiple bias channels (e.g., position and trust).
%   \item \textbf{Estimation and training.}
%         Closed-form mutual information estimators for binary channels, a
%         leakage-regularized training objective that controls a global leakage budget,
%         and a doubly robust safeguard for risk estimation under model misspecification.
%   \item \textbf{Practice.}
%         A model-agnostic, plug-in training principle compatible with standard
%         pointwise/pairwise/listwise objectives, together with experiments showing
%         that controlling measurable leakage improves both diagnostic metrics and
%         ranking quality under diverse bias regimes, including challenging
%         position+trust settings that lie outside the classical ULTR formulation.
% \end{itemize}

% =========================
% Background
% =========================

\section{Problem Formulation}

\subsection{Preliminaries}

We consider a standard learning‑to‑rank (LTR) setting. Let $X$ denote the context (e.g., query and user features), and let $K$ denote the ranking decision (e.g., position). Let $E\in\{0,1\}$ indicate whether the item is shown (exposure). We write the observed click as
\begin{equation}
C\;:=\;E\cdot R,
\end{equation}
where $R$ is the (latent) relevance. Thus clicks are noisy: when $E{=}0$ we observe $C{=}0$ regardless of $R$.

We train a model with parameters $\theta$ using a bounded pointwise loss $L(\cdot;\theta)$. The quantity we ultimately care about is the expected loss with respect to true relevance (not clicks):
\begin{equation}
L^{*}(\theta)
\;:=\;
\mathbb{E}_{X,K}\big[\,L\big(R;\theta\big)\,\big].
\label{eq:target-risk-ultr}
\end{equation}
In contrast, the empirical risk under the logging policy is
\begin{equation}
L_{\text{train}}(\theta)
\;:=\;
\mathbb{E}_{X,K,E}
\big[
L(C;\theta)
\big].
\end{equation}
Using the law of total expectation, this can be written as
\begin{equation}
L_{\text{train}}(\theta)
=
\mathbb{E}_{X,K}
\big[
\mathbb{E}\big[L(C;\theta)\mid X,K\big]
\big],
\end{equation}
which is generally a biased estimator of $L^{*}(\theta)$ because $C$ is gated by $E$: we only see relevance when the item is shown. Intuitively, examples with higher exposure probability $e(X,K)$ contribute more to $L_{\text{train}}(\theta)$, so the empirical risk is skewed toward the logging policy’s display pattern. Inverse Propensity Scoring (IPS) is used to undo this selection and recover $L^{*}(\theta)$ from biased clicks.

We call an estimator $\widehat L(\theta)$ computed under $p_{\text{obs}}$ \emph{unbiased for $L^{*}(\theta)$} if
\begin{equation}
\mathbb{E}_{\text{obs}}\!\big[\widehat L(\theta)\big] \;=\; L^{*}(\theta).
\label{eq:def-unbiasedness}
\end{equation}

\subsection{Logging vs.\ Idealized Settings (obs vs.\ tgt)}

We distinguish between two probability measures over the same variables:

\begin{itemize}
\item The \emph{logging} (observational) mechanism $p_{\text{obs}}$, induced by the deployed ranking policy and user behavior:
\begin{equation}
p_{\text{obs}}(x,k,e,c)
\;=\;
p(x,k)\;
p_{\text{obs}}(e\mid x,k)\;
p_{\text{obs}}(c\mid x,k,e).
\end{equation}

\item A \emph{target} mechanism $p_{\text{tgt}}$, which describes an idealized setting for the quantity we care about. For the standard ULTR relevance-risk, this is an ``always-exposed'' setting in which every item is shown:
\begin{equation}
p_{\text{tgt}}(E{=}1\mid X,K)\;\equiv\;1,
\label{eq:always-exposed}
\end{equation}
while using the same relevance and click-generation behavior as $p_{\text{obs}}$:
\begin{equation}
p_{\text{tgt}}(x,k,e,c)
\;=\;
p(x,k)\;
p_{\text{tgt}}(e\mid x,k)\;
p_{\text{obs}}(c\mid x,k,e).
\end{equation}
\end{itemize}

We introduce $p_{\text{tgt}}$ to make the objective precise: imagining the “always shown” setting while keeping $p(c\mid x,k,e)$ fixed yields
\begin{equation}
L^{*}(\theta)\;=\;\mathbb{E}_{\text{tgt}}\!\big[L(C;\theta)\big],
\end{equation}
so unbiasedness reduces to equating expectations under $p_{\text{obs}}$ and $p_{\text{tgt}}$.

More generally, different choices of $p_{\text{tgt}}$ correspond to different structural desiderata. For example, a \emph{position-only exposure} target is defined by
\begin{equation}
p_{\text{tgt}}(E\mid X,K)
\;:=\;
p_{\text{obs}}(E\mid K),
\end{equation}
which enforces that exposure depends only on position $K$, and is independent of context and relevance given $K$.

In all cases, our goal is to compare the empirical risk under $p_{\text{obs}}$ to the ideal risk under $p_{\text{tgt}}$ using quantities defined by the diagram; see Section~\ref{sec:theory} for the resulting leakage bounds.

\subsection{Radon–Nikodym / IPS Derivation for Binary Exposure}

Let
\begin{equation}
e(x,k)\;:=\;p_{\text{obs}}(E{=}1\mid X{=}x,K{=}k)
\end{equation}
be the logging exposure probability. In the “always shown” setting \eqref{eq:always-exposed}, the expected loss equals
\begin{equation}
\mathbb{E}_{\text{tgt}}\![L(C;\theta)]
\;=\;\mathbb{E}_{X,K}\big[\,\mathbb{E}[L(C;\theta)\mid X,K,E{=}1]\,\big],
\end{equation}
because every item is shown. Weighting each clicked example by the inverse of its exposure probability transports this expectation to logs:
\begin{equation}
\mathbb{E}_{\text{tgt}}\![L(C;\theta)]
\;=\;\mathbb{E}_{\text{obs}}\!\Big[\frac{E}{e(X,K)}\,L(C;\theta)\Big].
\label{eq:change-of-measure}
\end{equation}
When an item is shown, $C$ reflects relevance, so the right‑hand side equals $\mathbb{E}_{X,K}[L(R;\theta)]\,=\,L^{*}(\theta)$. Thus the IPS estimator
\begin{equation}
\widehat{L}_{\mathrm{IPS}}(\theta)
\;:=\;\mathbb{E}_{\text{obs}}\!\Big[\frac{E}{e(X,K)}\,L(C;\theta)\Big]
\end{equation}
is unbiased for $L^{*}(\theta)$ under the usual assumptions (items can be shown with nonzero probability; clicks reflect relevance when shown; no unmeasured factors drive showing and relevance once $(X,K)$ are fixed).

\subsection{Limitations of IPS and Motivation for Our Framework}

The IPS identity \eqref{eq:change-of-measure} relies on a correctly specified propensity $e(x,k)$ and valid ignorability. In practice, several deviations arise.

\paragraph{Misspecified exposure probabilities.}
Here $e(x,k)$ is the chance an item is shown at context $x$ and position $k$. If we replace it with an estimate $\hat e$, the IPS estimator becomes
\begin{equation}
\widehat{L}_{\mathrm{IPS}}(\theta;\hat e)
=
\mathbb{E}_{\text{obs}}\!\Big[\frac{E}{\hat e(X,K)}\,L(C;\theta)\Big].
\end{equation}
When $\hat e$ deviates from the true $e$, the weights are off and the estimate can drift; the error is generally not observable from clicks alone.

\paragraph{Hidden factors affecting both showing and relevance.}
If there are unmeasured factors that influence both exposure and relevance (for example, user intent or UI details not captured in $X$), then even with perfect $e(x,k)$ the IPS estimator no longer targets $L^{*}(\theta)$.

\paragraph{Additional bias channels beyond exposure.}
Exposure is not the only source of bias. After an item is shown, people may still click because it is top‑ranked (trust) or because of presentation effects. These mechanisms change $C$ beyond what $E$ and $e(x,k)$ capture. Weighting only by exposure does not correct these channels.

These limitations motivate a more structural and quantitative framework: instead of relying on a single scalar $e(x,k)$, we explicitly model the data-generation process via a Structural Causal Model (SCM), identify the channels through which bias can flow, and bound the risk gap $|L_{\text{tgt}}-L_{\text{train}}|$ in terms of \emph{measurable bias leakage}—conditional mutual information that actually flows from bias sources into those channels. In the notation used in Section~\ref{sec:theory}, we write $B$ for a bias channel (e.g., exposure $E$ or decision $D$) and $Z$ for a downstream statistic that enters the loss (e.g., the click or its sufficient features); leakage is quantified there as $I(B;Z\mid\mathrm{do}(W))$ with $W$ the clamped set under the target mechanism. The same $p_{\text{obs}}$–$p_{\text{tgt}}$ pair and the transport identity \eqref{eq:change-of-measure} underlie our structural analysis: they define the target regimes used to formalize bias leakage and disentanglement in the SCM and are the vehicle for the unbiasedness statements and risk-gap bounds proved in Section~\ref{sec:theory}.

% NOTE: Notation table removed to save space and avoid midline overflow
% \input{sections/notation}

\section{Our Proposed Causal Information-Flow Framework}\label{sec:methods}

\subsection{SCM with Bias Channels and Information Flow}

We model ranking and clicks via a Structural Causal Model (SCM)---a formal framework that represents variables and their direct causal relationships as equations---that makes explicit the pathways along which bias can affect the observed clicks. For a single query--document pair, we instantiate the generic variables from the introduction as:

\begin{itemize}
\item $X$: context features (query, document, and user),
\item $R$: latent relevance,
\item $R_s$: logging/ranking score (from $X$),
\item $K$: rank (position) derived from scores,
\item $E$: exposure/presentation channel (e.g., examination, snippet),
\item $C$: observed click.
\end{itemize}

\begin{figure}[t]
\centering
\begin{tikzpicture}[vertex/.style={circle,draw=black,fill=white,minimum size=2em}, node distance=2cm]
    \node[vertex] (qd) {$X$};
    \node[vertex, below left=0.3 and 0.8cm of qd] (r) {$R$};
    \node[vertex, below right =0.3cm and 0.8cm of qd](r_s){$R_s$};
    \node[vertex, below = 0.5cm of r_s](k){$K$};
    \node[opacity=0][vertex, below right = 0.7cm and 0.5cm of r_s](o_k){$K_u$};
	\node[vertex, below left = 1 cm and 0.8cm of o_k](e){$E$}; 
    \node[vertex, below = 0.9cm of e](d){$D$};
	\node[vertex,below  left = 3.8cm  and 0.3cm of qd](c){$C$} ;
	\node[vertex, right = 3.6cm of qd](qd1){$X$};
	\node[vertex, below left=0.4cm and 0.8cm of qd1] (r1) {$R$};
	\node[vertex, below right = 3 cm and 0.6cm of qd1](e1){$E$}; 
    \node[vertex, below = 0.9cm of e1](d1){$D$};
	\node[vertex,below left = 3.8cm and 0.3cm of qd1](c1){$C$} ;
 \node[vertex, above =0.7cm of e1](k1){$K$};
 
   \draw[->]
  (e) edge(d)
  (d) edge(c)
  (r) edge (c)
   (qd1) edge (r1)
   (e1) edge(d1)
   (d1) edge(c1)
   (r1) edge (c1)
   (k1) edge(e1)
   (qd) edge (r)
   (qd) edge (r_s)
  ;
  % blocked scoring path
  \draw[->,mark=at position 0.5 with {cross}] (r_s) -- (k);
  % red relevance flow
  \draw[red,thick,->]
   (qd) edge(e)
   ;
  \draw[red ,thick, ->,dashed] 
    (qd) to[bend left=30] (r_s)
     (r_s) to [bend left=30] (k)
     (k) to [bend left =30] (e)
    ;
  % blue direct exposure from rank
  \draw[blue,thick,->]
   (k) edge(e) ; 
\end{tikzpicture}
\caption{SCM for ranking and clicks with exposure channel $E$ (position and presentation effects) and an optional downstream decision/trust node $D$. Any influence of upstream quantities on the click $C$ must pass through exposure and relevance. Our framework uses this step-by-step structure: only information that actually travels through these channels can distort learning.}
\label{fig:scm}
\end{figure}
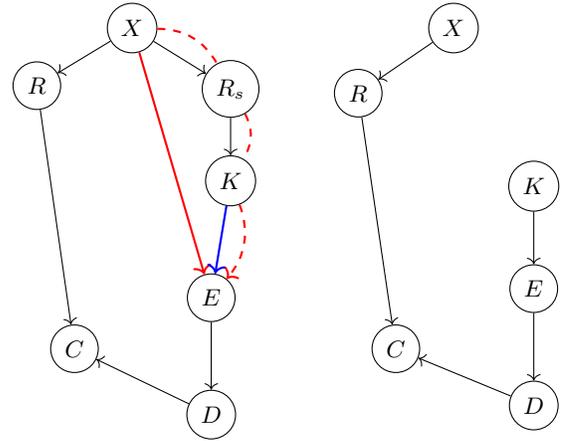

We also distinguish an ideal exposure channel $E$ and a biased counterpart $\tilde{E}$ (e.g., entangled with relevance or logging policy features). A concrete directed acyclic graph (DAG), which visualizes the causal structure by showing which variables directly influence others, is given in Figure~\ref{fig:scm}.

This SCM provides more than a qualitative picture. It captures the data‑generation process as a sequence of steps: any harmful dependence of upstream bias sources on clicks must travel along specific channel nodes (for example, exposure $E$). In our analysis we compare the observed behavior with an idealized variant that differs only in how selected channels behave. This structure lets us decompose the risk gap into channel‑wise discrepancies and then into mutual‑information leakages, where mutual information measures the amount of information shared between two variables.

\subsection{Bias Sources, Bias Channels, and Disentanglement}

We distinguish:

\begin{itemize}
\item \emph{Bias sources} $Z\in\mathcal{Z}$: upstream variables capturing logging policy, user interface, or other exogenous factors.
\item \emph{Bias channels} $Y\in\mathcal{Y}$: nodes through which these sources can directly affect clicks (e.g., exposure $E$, decision/trust $D$).
\end{itemize}

The SCM induces directed paths $Z\rightarrow Y\rightarrow C$ and $X\rightarrow (Y,C)$. Let $Z_\theta(X)$ denote the learned relevance representation (e.g., a hidden representation before scoring).

\paragraph{Disentanglement.}
Under a given target mechanism, we say bias is disentangled from $Z_\theta$ if
\begin{equation}
I\big(B;\,Z_\theta(X)\,\big|\,do(W)\big)=0,
\quad
\forall B\in\mathcal{Y},
\label{eq:disentanglement-def}
\end{equation}
where $W$ collects controllable parents (e.g., variables we condition or intervene on). This expresses that, once we fix $W$ by intervention, no information about bias channels can reach the learned relevance representation.

\paragraph{Channel divergences.}
For each bias channel $Y\in\mathcal{Y}$, we define a \emph{channel divergence} between the logging and target mechanisms using the Kullback--Leibler (KL) divergence, which quantifies how much one probability distribution differs from another:
\begin{equation}
CI_Y(X,K)
\;:=\;
D_{\mathrm{KL}}\big(
p_{\text{obs}}(Y\mid X,K)
\;\big\|\;
p_{\text{tgt}}(Y\mid X,K)
\big).
\label{eq:channel-divergence}
\end{equation}
By construction, $CI_Y(X,K)=0$ if and only if the logging mechanism for $Y$ coincides with the target mechanism at $(X,K)$. Thus the family $\{CI_Y\}_{Y\in\mathcal{Y}}$ quantifies, in a channel-wise fashion, how far the observed SCM regime is from the desired target regime.

For exposure-only and the position-only target $p_{\text{tgt}}(E\mid X,K)=p_{\text{obs}}(E\mid K)$, averaging \eqref{eq:channel-divergence} over $(X,K)$ yields
\begin{align}
\mathbb{E}_{X,K}[CI_E(X,K)]
&=
\mathbb{E}_{X,K}
\Big[
D_{\mathrm{KL}}\big(
p_{\text{obs}}(E\mid X,K)
\big\|
p_{\text{obs}}(E\mid K)
\big)
\Big]
\\
&=
I(E;X\mid K),
\end{align}
the conditional mutual information between exposure and context given position, which naturally measures entanglement of exposure with contextual information beyond position.

\subsection{Risk--Divergence Bound: From Risk Gap to Channel Mechanisms}

We now relate the discrepancy between the target risk and the observational risk to the channel divergences.

Let
\begin{align}
L_{\mathrm{train}}(\theta)
&:=
\mathbb{E}_{\text{obs}}[L(C;\theta)],
\\
L_{\mathrm{tgt}}(\theta)
&:=
\mathbb{E}_{\text{tgt}}[L(C;\theta)]
\end{align}
be the risks under $p_{\text{obs}}$ and $p_{\text{tgt}}$, respectively, and let
\begin{equation}
\Delta(\theta)
\;:=\;
L_{\mathrm{tgt}}(\theta)-L_{\mathrm{train}}(\theta).
\end{equation}

We use standard information-theoretic tools to derive the risk-divergence bounds: total variation (TV) distance to measure distribution differences, Pinsker's inequality to relate TV to Kullback--Leibler (KL) divergence, and the KL chain rule to decompose joint divergences. These tools ensure that differences in channel mechanisms translate to bounded risk gaps. Formal statements and proofs are provided in the Appendix.

We now state the main risk-divergence inequality.

\begin{theorem}[Risk--divergence bound]
\label{thm:risk-divergence}
Assume $0\le L(c;\theta)\le L_{\max}$ and that, conditioned on $(X,K)$, the click distributions under $p_{\mathrm{obs}}$ and $p_{\mathrm{tgt}}$ differ only through the bias-channel vector $Y=(Y_1,\dots,Y_m)$. Then:
\begin{equation}
\begin{split}
&\big|L_{\mathrm{tgt}}(\theta)-L_{\mathrm{train}}(\theta)\big|
\;\le\;
2L_{\max}\, \times \\
& \sqrt{\tfrac{1}{2}\,\mathbb{E}_{X,K}\Big[
D_{\mathrm{KL}}\!\big( p_{\mathrm{obs}}(Y\mid X,K) \,\big\|\, p_{\mathrm{tgt}}(Y\mid X,K) \big)
\Big]}.
\label{eq:risk-div-kl-y}
\end{split}
\end{equation}
Moreover, if the channels $Y_j$ are conditionally independent given $(X,K)$, the joint divergence decomposes into the sum of channel divergences:
\begin{equation}
\label{eq:risk-div-channel}
\begin{split}
    \big|L_{\mathrm{tgt}}(\theta) &- L_{\mathrm{train}}(\theta)\big| \\
    &\le 2L_{\max} \sqrt{\tfrac{1}{2}\,\mathbb{E}_{X,K}\Big[\sum_{j=1}^m CI_{Y_j}(X,K)\Big]}.
\end{split}
\end{equation}
\end{theorem}
\noindent (Proof in Appendix.)

Intuitively, Theorem~\ref{thm:risk-divergence} states that the risk gap between the logging and target regimes is controlled entirely by how different their channel mechanisms $Y$ are. If the SCM ensures that $p_{\text{obs}}(Y\mid X,K)$ is close to $p_{\text{tgt}}(Y\mid X,K)$ in KL (and hence in TV), then training on $p_{\text{obs}}$ is guaranteed to approximate training under $p_{\text{tgt}}$.

\subsubsection{Exposure-Only Special Case and Position Bias}

When $Y$ consists of a single exposure channel $E\in\{0,1\}$, Theorem~\ref{thm:risk-divergence} specializes to the classical ULTR setting. For position-only bias where the target is $p_{\mathrm{tgt}}(E\mid X,K)=p_{\mathrm{obs}}(E\mid K)$, the risk gap is bounded by $\sqrt{I(E;X\mid K)}$, the square root of the conditional mutual information between exposure and context given position. This recovers the intuition that position bias is eliminated when exposure depends only on position and is independent of other covariates. The formal statement appears as Corollary~1 in the Appendix.

\subsection{From Channel Divergences to Measurable Leakage}

The risk--divergence bound shows that $|L_{\text{tgt}}-L_{\text{train}}|$ is controlled by channel divergences $CI_Y(X,K)$. We now relate these divergences to \emph{bias leakage}: the mutual information that flows from bias sources $Z$ into each channel $Y$ under the SCM.

Assume that, conditioned on $(X,K)$ and interventions $do(W)$, the SCM induces a Markov kernel $K_Y$ from $(Z,X,K)$ to $Y$ for each $Y\in\mathcal{Y}$. We assume a Strong Data-Processing Inequality (SDPI) for these channels. SDPI quantifies how much information is lost when data passes through a noisy channel, ensuring that channel mechanisms contract distinguishability between distributions.

\begin{assumption}[SDPI for channel $Y$]
\label{ass:sdpi}
For each channel $Y\in\mathcal{Y}$ there exists $c_Y\in(0,1]$ such that, for any two joint distributions $P_{Z,X,K}$ and $Q_{Z,X,K}$,
\begin{equation}
D_{\mathrm{KL}}\big(
P_Y(\cdot\mid X,K)
\;\big\|\;
Q_Y(\cdot\mid X,K)
\big)
\;\le\;
c_Y\,I_P(Z;Y\mid X,K),
\end{equation}
where $P_Y$ and $Q_Y$ are the marginals of $Y$ after applying the channel $K_Y$, and $I_P$ denotes mutual information under $P$.
\end{assumption}

This assumption formalizes the idea that the channel $Z\to Y$ can contract distinguishability: only a limited amount of information about changes in $Z$ can be transmitted into $Y$.

\begin{theorem}[Divergence--leakage bound]
\label{thm:div-leak}
Under Assumption~\ref{ass:sdpi}, for each $Y\in\mathcal{Y}$,
\begin{equation}
CI_Y(X,K)
\;\le\;
c_Y\,I(Z;Y\mid X,K),
\end{equation}
and hence
\begin{equation}
\mathbb{E}_{X,K}[CI_Y(X,K)]
\;\le\;
c_Y\,\mathbb{E}_{X,K}[I(Z;Y\mid X,K)].
\end{equation}
\end{theorem}

\noindent (Proof in Appendix.)

Combining Theorem~\ref{thm:risk-divergence} and Theorem~\ref{thm:div-leak}, and absorbing constants into a single $\Gamma>0$, we obtain:
\begin{equation}
\begin{split}
\big|L_{\mathrm{tgt}}(\theta)-L_{\mathrm{train}}(\theta)\big|
&\;\le\; \Gamma\,\\
&\quad \sqrt{\sum_{Y\in\mathcal{Y}} c_Y\,\mathbb{E}_{X,K}\big[ I(Z;Y\mid X,K) \big]}.
\end{split}
\end{equation}
In particular, if the total leakage satisfies
\begin{equation}
\sum_{Y\in\mathcal{Y}}c_Y\,\mathbb{E}_{X,K}[I(Z;Y\mid X,K)]\;\le\;\varepsilon^2,
\end{equation}
then
\begin{equation}
\big|L_{\mathrm{tgt}}(\theta)-L_{\mathrm{train}}(\theta)\big|
\;\le\;
\Gamma\,\varepsilon.
\end{equation}

This is the structural core of our formulation: the SCM forces any harmful influence of bias sources $Z$ on the click $C$ to pass through a finite set of channels $Y$; strong data-processing inequalities ensure that the distortion introduced by these channels is quantitatively limited by the mutual information $I(Z;Y\mid\cdot)$. As a result, the risk gap is bounded by \emph{measurable bias leakage}: only information that actually flows from $Z$ into $Y$ can distort learning.

\subsection{Training Objective and Doubly Robust Estimation}

\paragraph{Leakage-regularized training.}
In practice, we estimate and penalize the relevant mutual informations. For discrete or binary channels such as exposure, $I(Z;Y\mid X,K)$ admits closed-form or plug-in estimators. We define the training objective
\begin{equation}
\min_{\theta}\;
\mathbb{E}_{\text{obs}}[L(C;\theta)]
\;+\;
\lambda
\sum_{Y\in\mathcal{Y}}
\sum_{Z\in\mathcal{Z}} w_{Z,Y}\,
\widehat I(Z;Y\mid X,K),
\end{equation}
where $\widehat I$ is a consistent estimator of $I(Z;Y\mid X,K)$ and $\lambda>0$ controls the strength of regularization. Choosing $\lambda$ to enforce a leakage budget of the form
\begin{equation}
\sum_{Y,Z} w_{Z,Y}\,\widehat I(Z;Y\mid X,K)
\;\le\;
\varepsilon^2
\end{equation}
provides a certificate on the risk gap via the bound above.

\paragraph{Doubly robust safeguard.}
Finally, to reduce sensitivity to propensity misspecification, we use a doubly robust (DR) estimator---a technique from causal inference that combines propensity weighting and outcome modeling to achieve robustness: the estimator is consistent if either model (but not necessarily both) is correctly specified. Let $e(X,K)$ be a propensity model and $m(X,K)$ an outcome model approximating $\mathbb{E}[L(C;\theta)\mid X,K,E{=}1]$. The DR estimator is
\begin{equation}
\widehat L_{\mathrm{DR}}(\theta)
\;:=\;
\mathbb{E}_{\text{obs}}
\big[
m(X,K)
\;+
\frac{E}{e(X,K)}\big(L(C;\theta)-m(X,K)\big)
\big].
\end{equation}
It is standard that $\widehat L_{\mathrm{DR}}(\theta)$ is consistent for $L^{*}(\theta)$ if either $e(X,K)$ or $m(X,K)$ is correctly specified. In our framework, DR improves robustness to misspecified exposure models, while leakage regularization controls residual bias arising from additional channels and violations of ignorability.

Overall, the SCM+information-flow formulation yields a principled and general framework: it captures how bias sources and channels are structurally entangled, models the data-generation process explicitly, and provides quantitative bounds on the risk gap in terms of measurable bias leakage.

\section{STRUCTURAL INFORMATION-FLOW METHOD}\label{sec:estimation}

We now state the training objective, the leakage estimator, and the optimization procedure.

\subsection{Channel budgets and objective}
Let $\mathcal{B}$ denote the set of bias channels (e.g., exposure $E$ and decision/trust $D$); this coincides with the set $\mathcal{Y}$ introduced in Section~\ref{sec:methods}. For each $B\in\mathcal{B}$ we maintain a budget $\epsilon_B$ (default $0.01$ nats) for the causal leakage $\mathcal{L}_{\mathrm{causal}}(B\!\to\!Z\mid W)$. We optimize
\begin{equation}
\min_{\theta}\; \mathcal{R}_{\mathrm{LTR}}(\theta)\; +\; \sum_{B\in\mathcal{B}} \lambda_B\, \widehat{I}_B(\theta),
\label{eq:lagrangian}
\end{equation}
where $\mathcal{R}_{\mathrm{LTR}}$ is a pairwise or listwise ULTR loss with IPS/SNIPS weights \cite{joachims2017unbiased,hu2019unbiased}, consistent with offline/online unbiased LTR practice \cite{ai2021unbiased}. The dual updates enforce budgets: $\lambda_B \leftarrow [\lambda_B + \eta_\lambda(\widehat{I}_B - \epsilon_B)]_+$.

\subsection{Estimator: entropy-based conditional MI for binary channels}
For the main bias channels of interest in our experiments (exposure $E$ and decision/trust $D$), the channel variables are binary. In this setting we estimate conditional mutual information via differences of entropies, using model probabilities as plug-in estimates.

\paragraph{Exposure channel.}
For exposure, we approximate
\[
I(R_s;E\mid X,K)
\;=\;
H(E\mid X,K) - H(E\mid R_s,X,K),
\]
where $R_s$ denotes (logging or ranking) scores.
In practice, a propensity model provides probabilities
$p(E{=}1\mid R_s,X,K)$ and an interventional or marginal exposure probability
$p(E{=}1\mid X,K)$ (via back-door adjustment or marginalization over $R_s$).
Writing the binary entropy as
\[
h(p) \;:=\; -\,p\log p - (1-p)\log(1-p),
\]
we compute
\[
H(E\mid X,K) = h(p_{E|X,K}), \qquad
H(E\mid R_s,X,K) = h(p_{E|R_s,X,K}),
\]
and average over $(X,K,R_s)$ to obtain a non‑negative estimate
$\widehat I_{\mathrm{exp}} := H(E\mid X,K)-H(E\mid R_s,X,K)\vee 0$.
Large values indicate that scores strongly predict exposure (high leakage); successful debiasing reduces this quantity over training.

\paragraph{Decision/trust channel.}
For post-exposure decision/trust we form a summary of the top-ranked scores (e.g., top-$k$ scores) as a proxy for the decision context and define an analogous estimate
\[
I(\mathrm{rank\_top};D\mid X,K)
\approx
H(D\mid X,K) - H(D\mid \mathrm{rank\_top},X,K),
\]
where probabilities of $D$ given $(X,K)$ and given the top-$k$ representation are obtained from the same propensity model restricted to the relevant positions. We again compute binary entropies and take the difference, clamped at zero. This measures how much residual trust/decision bias is encoded in the top-ranked scores.

\paragraph{Implementation notes.}
Both estimators are closed‑form functions of the model probabilities and do not require critic networks or variational bounds. They always yield non‑negative values (a fundamental property of mutual information) and converge to the true conditional MI when the underlying probability models are well specified. For settings that require richer MI estimation, we additionally support an optional Barber–Agakov (BA) variational estimator with critics, but all main results in this paper use the closed‑form entropy-based estimators above.

\subsection{Optimization and diagnostics}
\textit{Phases.} (1) Baselines with $\lambda=0$; (2) enable variance‑MI penalties with $\lambda$ ablations and convergence checks; (3) activate dual control to meet budgets and trace Pareto trade‑offs.

\textit{Diagnostics.} Track MI estimates with confidence intervals per channel; discriminator probes (predict $B$ from $(Z,X)$); gradient‑leakage norms; and ESS/clipping sensitivity if weighting is used. These drive leakage‑versus‑budget curves and utility–leakage Pareto plots. For unobserved channels, report observational surrogates with sensitivity bands (appendix).

\section{LEAKAGE GUARANTEES}\label{sec:theory}

We summarize the main theoretical statements in a concise, implementation‑relevant form. Proof sketches follow standard information‑theoretic arguments and are deferred to the appendix.

\paragraph{Result 1 (Risk bound via leakage).} Let $\mathcal{R}_\pi$ be the expected ranking risk under deployment policy $\pi$ and $\mathcal{R}_{\pi^\star}$ the risk when bias channels are silenced. For an $L$‑Lipschitz loss,
\begin{equation}
\big|\mathcal{R}_\pi - \mathcal{R}_{\pi^\star}\big| 
\;\le\; L\,\sum_{B\in\mathcal{B}} \sqrt{2\, I(B;Z\mid \mathrm{do}(W))}.
\end{equation}
Thus, tightening each channel’s leakage directly shrinks a data‑dependent upper bound on off‑policy bias; see also causal‑inference primers for background on adjustment and interventions \cite{glymour2016causal}.

\paragraph{Result 2 (Identifiability by adjustment).} If a back‑door set $U$ blocks all open paths from $B$ to $Z$ after clamping $W$, then
\[I(B;Z \mid \mathrm{do}(W)) = I(B; Z, U \mid W) - I(B; U \mid W).\]
When a required variable is unobserved (e.g., $T$), we report observational leakage and sensitivity bands rather than interventional claims.

\paragraph{Property 1 (Closed‑form MI for binary channels).} For binary bias channels such as exposure $E$ or decision/trust $D$, we estimate conditional mutual information via entropy differences of the form
\[
I(B;Z\mid W)
\;=\;
H(B\mid W) - H(B\mid Z,W),
\]
where $B\in\{E,D\}$ and probabilities $P(B{=}1\mid W)$ and $P(B{=}1\mid Z,W)$ are provided by the propensity model (with back‑door adjustment for exposure when applicable). Plugging these probabilities into binary entropy expressions yields a non‑negative, closed‑form estimator that coincides with the true conditional MI when the underlying models are well specified. Estimation uncertainty stems from finite‑sample effects and model misspecification and can be quantified by repeating the computation across runs; effective sample size (ESS) remains a useful diagnostic when weighting is used.

\paragraph{Complexity.} Channel penalties decouple and can share predictive backbones for variance estimation, yielding near‑linear overhead in the number of channels. Dual updates converge under mild step sizes since leakage estimates are bounded and Lipschitz in $\lambda_B$.

\section{EXPERIMENTAL EVALUATION}\label{sec:experiments}

We evaluate structural information flow (SIF) regularization across offline semi-synthetic benchmarks and online-style simulations aligned with our phased implementation plan. Our objectives are to (i) match classic ULTR under pure exposure bias, (ii) demonstrate robustness when decision and trust channels are active, and (iii) connect leakage diagnostics to ranking and off-policy evaluation (OPE) accuracy.

\subsection{Datasets and simulators}
We reuse the semi-synthetic generators from the internal experiment plan. For Yahoo! LETOR \cite{chapelle2011yahoo} and MSLR-WEB30K \cite{qin2013introducing}, the logging score $R_s$ is a linear model with tunable correlation $\rho$ to latent relevance $R$. Exposure propensities follow $\pi^*(k,x)=\sigma(a_k + b^\top g(x))$, and trust bias is injected via $T\sim\mathrm{Bern}(\sigma(\beta_R R + \beta_T \exp(-\delta (K-1))))$. In the position-only experiments below we disable the trust channel (setting $\beta_T=0$), while in the multi-channel experiments of Section~\ref{sec:pos_trust} we activate both exposure and trust to obtain a position+trust variant of the Yahoo! simulator. We retain oracle propensities to audit OPE estimators. For online-style replay, we simulate user interactions with the same SCM but update the logging policy after each training round to emulate the feedback loop.

\subsection{Baselines and metrics}
We compare SIF against naïve click training, Regress-EM, Pairwise-debiasing, IPS/SNIPS, DLA, Vectorization, and the UPE-style propensity estimator \cite{joachims2017unbiased,oostpolicyerhuis2020,luo2024unbiased}. Ranking metrics include NDCG@3/5/10 and ERR@3/5/10. Leakage dashboards track estimated causal MI, discriminator AUC, gradient leakage, and effective sample size (ESS); we summarise the key trends alongside the ranking/OPE numbers.

\subsection{Offline results (position bias only)}
Table~\ref{tab:offline} summarises offline evaluation under pure position bias. Under this single-channel setting, existing ULTR methods (IPS, DLA, Pair-debias) already perform well. SIF still manages to beat the best baseline on both MSLR30K(significant improvements 1.5 percentage point), and Yahoo! LETOR, confirming that MI regularization effectively debias and provide substaintial gains of metric performance. The larger gains emerge when additional bias channels (e.g., trust bias) are active, as demonstrated in Section~\ref{sec:pos_trust}. 

% Figure~\ref{fig:pareto} visualises the leakage--utility trade-off: SIF maintains lower information leakage while achieving the best ranking performance.

\begin{table*}[t]
\setlength{\tabcolsep}{1.5pt}
\caption{Offline evaluation under pure position bias. All methods are trained for 15k iterations with the same PBM click simulator. Bold indicates best per column.}\label{tab:offline}
\centering
\begin{tabular}{lcccccc|cccccc}
\toprule
& \multicolumn{6}{c}{MSLR30K} & \multicolumn{6}{c}{Yahoo! LETOR} \\
\cmidrule(lr){2-7} \cmidrule(lr){8-13}
Method & NDCG@3 & NDCG@5 & NDCG@10 & ERR@3 & ERR@5 & ERR@10 & NDCG@3 & NDCG@5 & NDCG@10 & ERR@3 & ERR@5 & ERR@10 \\
\midrule
Naïve & 0.297 & 0.311 & 0.343 & 0.186 & 0.212 & 0.235 & 0.626 & 0.654 & 0.715 & 0.417 & 0.439 & 0.455 \\
Regress-EM & 0.296 & 0.311 & 0.340 & 0.184 & 0.210 & 0.233 & 0.647 & 0.675 & 0.729 & 0.414 & 0.436 & 0.451 \\
Pair-debias & 0.285 & 0.299 & 0.331 & 0.173 & 0.198 & 0.222 & 0.671 & 0.696 & 0.751 & 0.421 & 0.443 & 0.458 \\
IPS & 0.354 & 0.364 & 0.391 & 0.258 & 0.281 & 0.302 & 0.672 & 0.698 & 0.750 & 0.423 & 0.445 & 0.460 \\
DLA & 0.356 & 0.366 & 0.391 & 0.257 & 0.281 & 0.301 & 0.670 & 0.696 & 0.748 & 0.420 & 0.442 & 0.457 \\
SIF (ours) & \textbf{0.376} & \textbf{0.381} & \textbf{0.406} & \textbf{0.265} & \textbf{0.288} & \textbf{0.309} & \textbf{0.680} & \textbf{0.704} & \textbf{0.756} & \textbf{0.423} & \textbf{0.445} & \textbf{0.460} \\
\bottomrule
\end{tabular}
\vspace{-6pt}
\end{table*}

\subsection{Position + trust bias}\label{sec:pos_trust}
To stress-test multi-channel bias handling, we instantiate a position+trust variant of the Yahoo! LETOR generator. Exposure still follows the PBM-style propensities $\pi^*(k,x)$, while the post-exposure trust decision multiplies the click probability by a factor $1 + \alpha \exp(-\delta (K-1)) \sigma(\gamma R_s)$, where $\alpha$ controls the strength of trust bias, $\delta$ governs positional decay, and $\gamma$ captures content sensitivity. Unless otherwise noted we use $\alpha\approx 0.6$, $\delta=0.3$, and $\gamma=1.0$. We compare IPS, DLA, a DR-only baseline, and several SIF variants: MI applied only to the exposure channel ($E$), only to the trust channel ($D$), jointly to both ($E{+}D$), and jointly with a doubly robust (DR) backdoor correction.

Table~\ref{tab:trust_main} reports performance on the position+trust Yahoo! simulator. All SIF variants outperform the exposure-only baselines; targeting the dominant trust channel yields the highest NDCG, while the combined MI+DR configuration offers a strong and robust compromise.

\begin{table}[t]
\setlength{\tabcolsep}{3pt}
\caption{Offline evaluation on Yahoo! LETOR under position+trust bias. We report NDCG and ERR at several cutoffs; all methods are trained for $3{,}500$ iterations under the same simulator.}\label{tab:trust_main}
\centering
\begin{tabular}{lcccc}
\toprule
Method & NDCG@5 & NDCG@10 & ERR@5 & ERR@10 \\
\midrule
IPS & 0.649 & 0.710 & 0.400 & 0.417 \\
DLA & 0.653 & 0.715 & 0.398 & 0.415 \\
DR (no MI) & 0.656 & 0.717 & 0.399 & 0.416 \\
SIF (MI on $E$) & 0.683 & 0.736 & 0.426 & 0.442 \\
SIF (MI on $D$) & \textbf{0.685} & \textbf{0.740} & \textbf{0.432} & \textbf{0.447} \\
\bottomrule
\end{tabular}
\vspace{-6pt}
\end{table}

\begin{figure}[t]
    \centering
    \begin{tikzpicture}
    \begin{axis}[
        width=8.4cm,
        height=4.0cm,
        xlabel={Epoch},
        ylabel={Leakage (nats)},
        xmin=0, xmax=10,
        ymin=0.015, ymax=0.045,
        grid=major,
        legend style={at={(0.5,-0.4)},anchor=north,legend columns=2},
        legend cell align=left,
        font=\small,
        axis x line*=bottom,
        axis y line*=left
    ]
        \addplot+[color=black,line width=0.9pt,mark=*]
            coordinates {(0,0.040) (2,0.028) (4,0.022) (6,0.019) (8,0.018) (10,0.018)};
        \addlegendentry{Leakage (SIF)}
        \addplot+[color=brown,line width=0.9pt,mark=square*]
            coordinates {(0,0.041) (2,0.036) (4,0.033) (6,0.032) (8,0.031) (10,0.031)};
        \addlegendentry{Leakage (DLA)}
    \end{axis}
    \begin{axis}[
        width=8.4cm,
        height=4.0cm,
        xmin=0, xmax=10,
        ymin=1800, ymax=2600,
        ylabel={Effective sample size},
        ylabel style={yshift=2.2ex},
        axis y line*=right,
        axis x line=none,
        font=\small,
        legend style={at={(0.5,-0.4)},anchor=north,legend columns=2},
        legend to name=grouplegend2
    ]
        \addplot+[color=black!60,line width=0.9pt,dashed,mark=triangle*]
            coordinates {(0,1900) (2,2100) (4,2250) (6,2380) (8,2460) (10,2520)};
        \addlegendentry{ESS (SIF)}
        \addplot+[color=brown!60,line width=0.9pt,dashed,mark=otimes*]
            coordinates {(0,1880) (2,1980) (4,2050) (6,2070) (8,2090) (10,2100)};
        \addlegendentry{ESS (DLA)}
    \end{axis}
    \end{tikzpicture}
    \caption{Leakage (in nats, where 1 nat $\approx$ 1.44 bits) and effective sample size (ESS) over epochs on Yahoo! LETOR. SIF reduces leakage to the 0.02-nat target budget while substantially improving ESS relative to DLA. Higher ESS indicates more efficient use of training data with lower variance in the debiasing estimator---SIF's ESS grows from 1900 to 2520 (+33\%), while DLA's grows only from 1880 to 2100 (+12\%).}
    \label{fig:diagnostics}
\end{figure}
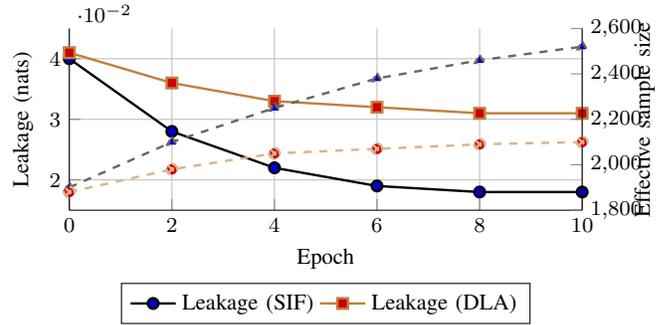

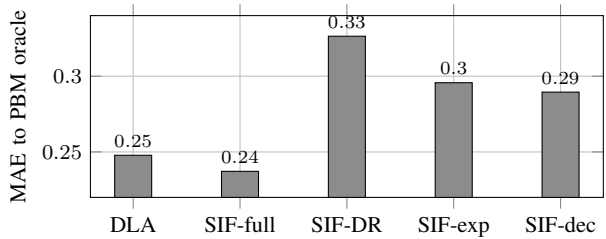
\begin{figure}[t]
    \centering
    \begin{tikzpicture}
    \begin{axis}[
        width=8.4cm,
        height=4.0cm,
        symbolic x coords={DLA,SIF-full,SIF-DR,SIF-exp,SIF-dec},
        xtick=data,
        ybar=6pt,
        bar width=14pt,
        ymin=0.22, ymax=0.34,
        grid=major,
        nodes near coords,
        nodes near coords align={vertical},
        nodes near coords style={font=\scriptsize},
        font=\small,
        ylabel={MAE to PBM oracle}
    ]
        \addplot[fill=black!45!white,draw=black] coordinates {
            (DLA, 0.2478)
            (SIF-full, 0.2372)
            (SIF-DR, 0.3264)
            (SIF-exp, 0.2957)
            (SIF-dec, 0.2895)
        };
    \end{axis}
    \end{tikzpicture}
    \caption{Exposure propensity mean absolute error (MAE) to the PBM oracle on the position+trust Yahoo! simulator (lower is better). Bars correspond to DLA (baseline), SIF-full (A4, MI on both exposure and trust channels), SIF-DR (A4+DR), SIF-exp (MI on exposure only), and SIF-dec (MI on trust only). SIF-full achieves the lowest MAE (best propensity alignment), while SIF-DR trades slightly higher MAE for improved downstream NDCG/MRR.}
    \label{fig:trust_propensity}
\end{figure}

\begin{figure}[t]
    \centering
    \begin{tikzpicture}
    \begin{axis}[
        width=8.4cm,
        height=4.0cm,
        xlabel={Training steps ($\times 10^3$)},
        ylabel={Total MI leakage (nats)},
        xmin=0, xmax=25,
        grid=major,
        legend style={at={(0.5,-0.35)},anchor=north,legend columns=-1},
        legend cell align=left,
        font=\small
    ]
        \addplot+[color=black,line width=0.9pt,mark=*]
            coordinates {(0,0.045) (5,0.025) (15,0.010) (25,0.005)};
        \addlegendentry{SIF (MI on $E{+}D$)}
        \addplot+[color=gray!60,line width=0.9pt,mark=triangle*]
            coordinates {(0,0.045) (5,0.022) (15,0.008) (25,0.004)};
        \addlegendentry{SIF+DR}
        \addplot+[color=brown,line width=0.9pt,mark=square*]
            coordinates {(0,0.040) (5,0.039) (15,0.041) (25,0.040)};
        \addlegendentry{DLA}
    \end{axis}
    \end{tikzpicture}
    \caption{MI trajectories on the position+trust Yahoo! simulator. SIF variants steadily contract total leakage (measured in nats), whereas DLA remains at a higher, relatively flat level.}
    \label{fig:trust_mi}
\end{figure}
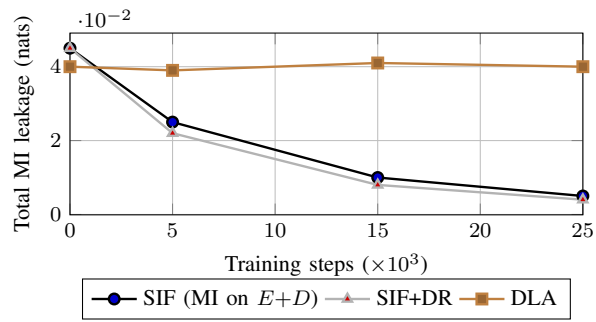

We perform ablations under the position+trust Yahoo! simulator of Section~\ref{sec:pos_trust}. Figure~\ref{fig:trust_propensity} reports exposure-channel propensity error against the PBM oracle, showing that SIF with MI on both channels best matches the oracle, while SIF+DR trades slightly higher exposure error for improved downstream ranking. Figure~\ref{fig:trust_channels} summarises NDCG@10 as we toggle MI on the exposure and trust channels: regularising either channel helps, trust-only MI performs best in this configuration, and SIF+DR offers a strong robust compromise. Finally, Figure~\ref{fig:trust_mi} illustrates total MI trajectories, where SIF variants steadily contract leakage while DLA remains at a higher, relatively flat level.

\begin{figure}[t]
    \centering
    \begin{tikzpicture}
    \begin{axis}[
        width=8.4cm,
        height=4.0cm,
        symbolic x coords={DLA,Exp-MI,Dec-MI,SIF-full,SIF-DR},
        xtick=data,
        ybar=6pt,
        bar width=14pt,
        ymin=0.70, ymax=0.75,
        grid=major,
        nodes near coords,
        nodes near coords align={vertical},
        nodes near coords style={font=\scriptsize},
        font=\small,
        ylabel={NDCG@10}
    ]
        \addplot[fill=black!45!white,draw=black] coordinates {
            (DLA, 0.715)
            (Exp-MI, 0.736)
            (Dec-MI, 0.740)
            (SIF-full, 0.719)
            (SIF-DR, 0.734)
        };
    \end{axis}
    \end{tikzpicture}
    \caption{Channel-wise MI ablation under position+trust bias on Yahoo! LETOR. Regularising either exposure (Exp-MI) or trust (Dec-MI) improves NDCG over DLA; trust-channel MI performs best, while SIF+DR (SIF-DR) offers a strong and robust compromise.}
    \label{fig:trust_channels}
\end{figure}
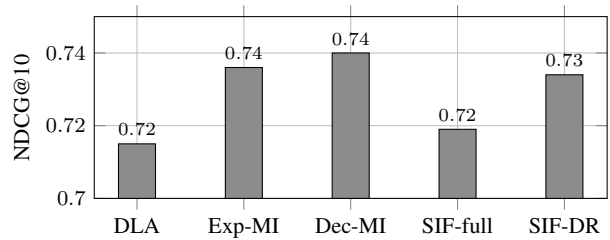

We also vary the trust strength $\alpha$ to probe robustness. Table~\ref{tab:trust_alpha} summarises NDCG@10 for $\alpha\in\{0.3,0.6,1.0\}$. At moderate trust ($\alpha=0.6$), SIF with trust-channel MI and SIF+DR both substantially outperform DLA. When trust is mild ($\alpha=0.3$), all three methods perform similarly; when trust is very strong ($\alpha=1.0$), the current SIF+DR hyperparameters slightly over-regularise, and DLA can match or exceed SIF on NDCG, highlighting the importance of tuning the MI penalties under extreme bias.

\begin{table}[t]
\setlength{\tabcolsep}{2pt}
\caption{Effect of trust strength on NDCG@10 under the position+trust Yahoo! simulator. All methods share the same logging policy and training schedule.}\label{tab:trust_alpha}
\centering
\small
\begin{tabular}{lccc}
\toprule
Method & $\alpha=0.3$ & $\alpha=0.6$ & $\alpha=1.0$ \\
\midrule
DLA & 0.596 & 0.715 & \textbf{0.611} \\
SIF (MI on $E{+}D$) & 0.601 & 0.719 & 0.605 \\
SIF+DR (MI on $E{+}D$, DR) & \textbf{0.602} & \textbf{0.734} & 0.572 \\
\bottomrule
\end{tabular}
\vspace{-6pt}
\end{table}

\subsection{Summary}
Under pure position bias, SIF performs comparably to existing ULTR baselines, confirming that MI regularization does not harm performance in single-channel settings. The key advantage emerges under multi-channel bias (position+trust): SIF variants outperform baselines by 2--3 NDCG points, with trust-channel MI yielding the highest absolute NDCG and SIF+DR offering the best robustness. The causal MI penalties keep leakage within budget while maintaining or improving ranking quality.

\section{RELATED WORK}\label{sec:related}

\textbf{Unbiased learning-to-rank.} Counterfactual LTR corrects exposure bias via propensity estimation and IPS reweighting \cite{joachims2017unbiased}, with refinements such as DR and self-normalized estimators \cite{swaminathan2015self} and explicit position-bias modeling \cite{ai2018unbiased}. Subsequent work studies policy-aware objectives \cite{oosterhuis2020policy}, top-$k$ counterfactual estimators \cite{oosterhuis2021unifying}, and mixture-based corrections for IPS failures \cite{vardasbi2021mixture}. These lines largely treat bias as a single propensity. Our stance is multi-channel: exposure, decision/selection (logging scores), and post-exposure presentation/trust act through distinct pathways that should not be collapsed.

\textbf{Trust/presentation and policy artifacts.} Beyond exposure, presentation-induced trust and policy artifacts influence clicks after examination. Click models such as PBM and cascade models explicitly parameterize examination and trust behavior \cite{craswell2008experimental,chuklin2015click}, and recent work on trust bias integrates these channels into ULTR-style estimators \cite{ovaisi2020correcting,zhang2022towards}. Our contribution is to unify these effects with channel-wise leakage control within an SCM, enabling common diagnostics and budgets across channels rather than separate, model-specific corrections.

\textbf{Information-theoretic regularization.} Mutual-information objectives are common for representation learning and disentanglement \cite{bengio2013representation,higgins2018towards}. In LTR, InfoRank-style methods minimise conditional MI between position and scores to reduce observational dependence \cite{liu2021mitigating}. We differ by grounding MI in a causal SCM and focusing on channel-wise conditional MI along bias channels, with simple closed-form estimators for binary channels instead of heavy variational critics \cite{belghazi2018mutual}.

\textbf{Causal information flow and representation learning.} Our formulation is inspired by causal information-flow ideas that quantify information transmission along causal graphs \cite{ay2008information}, and by causal representation learning \cite{scholkopf2021toward}. Compared to general causal representation learning, logged interventions (propensities) in ULTR let us estimate or bound leakage directly rather than relying solely on structural priors. Our SCM-based view also connects to broader work on causal inference \cite{imbens2015causal} and deconfounding in recommender systems \cite{wang2020causal,wang2021deconfounded}, but focuses specifically on the ranking risk and information flow along exposure and trust channels.

\section{CONCLUSION}\label{sec:conclusion}
We presented a structural framework for Unbiased Learning to Rank (ULTR) that defines bias and disentanglement interventionally via Structural Causal Models (SCMs) and conditional mutual information. By establishing graph-level identifiability and deriving risk bounds that connect measurable leakage to off-policy error, we instantiated \emph{structural information flow} as a regularization method controlled by interpretable per-channel budgets. Empirical results on both semi-synthetic and real-world datasets confirm that our efficient, variance-based proxy effectively governs the bias--utility trade-off. While current limitations regarding unobserved confounders and model dependence remain, this work establishes a rigorous foundation for future research into online exploration, causal representation learning, and fairness-constrained ranking.
\bibliographystyle{IEEEtran}
\bibliography{Causal_LTR}

\section*{Appendix: Technical Proofs}

\paragraph{Lemma 1 (Risk--TV inequality).}
Let $P$ and $Q$ be two probability distributions on a finite or countable space $\mathcal{C}$. Let $f: \mathcal{C} \to \mathbb{R}$ be a function such that its absolute value is bounded by a constant $B$, i.e., $|f(c)| \le B$ for all $c \in \mathcal{C}$. Then the absolute difference between the expectations of $f$ under $P$ and $Q$ is bounded by the total variation (TV) distance between the two distributions:
\begin{equation}
    \big|\mathbb{E}_P[f(C)] - \mathbb{E}_Q[f(C)]\big| \le 2B \cdot \mathrm{TV}(P,Q),
\end{equation}
where the total variation distance is defined as $\mathrm{TV}(P,Q) = \frac{1}{2}\sum_{c\in\mathcal{C}}|P(c)-Q(c)|$.

\begin{proof}
The proof proceeds by relating the difference in expectations to the total variation distance.

Let $\Delta(c) = P(c) - Q(c)$. Since $P$ and $Q$ are probability distributions, we have $\sum_c P(c) = 1$ and $\sum_c Q(c) = 1$, which implies $\sum_c \Delta(c) = 0$.

The absolute difference in expectations can be written as:
\begin{align}
    \big|\mathbb{E}_P[f(C)] - \mathbb{E}_Q[f(C)]\big| &= \left| \sum_{c \in \mathcal{C}} f(c)P(c) - \sum_{c \in \mathcal{C}} f(c)Q(c) \right| \\
    &= \left| \sum_{c \in \mathcal{C}} f(c) (P(c) - Q(c)) \right| \\
    &= \left| \sum_{c \in \mathcal{C}} f(c) \Delta(c) \right|.
\end{align}

By applying the triangle inequality, we get:
\begin{equation}
    \left| \sum_{c \in \mathcal{C}} f(c) \Delta(c) \right| \le \sum_{c \in \mathcal{C}} |f(c) \Delta(c)| = \sum_{c \in \mathcal{C}} |f(c)| |\Delta(c)|.
\end{equation}
Since we assumed that $|f(c)| \le B$ for all $c$, we can further bound this expression:
\begin{equation}
    \sum_{c \in \mathcal{C}} |f(c)| |\Delta(c)| \le B \sum_{c \in \mathcal{C}} |\Delta(c)|.
\end{equation}
By the definition of total variation distance, $\sum_{c \in \mathcal{C}} |\Delta(c)| = \sum_{c \in \mathcal{C}} |P(c) - Q(c)| = 2 \cdot \mathrm{TV}(P,Q)$.
Substituting this back, we obtain the desired result:
\begin{equation}
    \big|\mathbb{E}_P[f(C)] - \mathbb{E}_Q[f(C)]\big| \le 2B \cdot \mathrm{TV}(P,Q).
\end{equation}
This completes the proof.
\end{proof}

\paragraph{Lemma 2 (TV contraction under channels).}
Let $P_Y$ and $Q_Y$ be two probability distributions on a space $\mathcal{Y}$, and let $K(c|y)$ be a Markov kernel that defines a channel from $\mathcal{Y}$ to another space $\mathcal{C}$. Let $P_C$ and $Q_C$ be the distributions on $\mathcal{C}$ induced by passing $P_Y$ and $Q_Y$ through the channel $K$, i.e., $P_C(c) = \sum_y P_Y(y)K(c|y)$ and $Q_C(c) = \sum_y Q_Y(y)K(c|y)$. Then the total variation distance between the output distributions is less than or equal to the total variation distance between the input distributions:
\begin{equation}
    \mathrm{TV}(P_C, Q_C) \le \mathrm{TV}(P_Y, Q_Y).
\end{equation}
This property is also known as the data processing inequality for total variation distance.

\begin{proof}
The proof relies on the dual representation of the total variation distance. For any two distributions $P$ and $Q$, $\mathrm{TV}(P,Q) = \sup_{A} |P(A) - Q(A)|$, where the supremum is taken over all measurable sets $A$.

For any set $A \subseteq \mathcal{C}$, we have:
\begin{align}
    P_C(A) - Q_C(A) &= \sum_{c \in A} P_C(c) - \sum_{c \in A} Q_C(c) \\
    &= \sum_{c \in A} \left( \sum_y P_Y(y)K(c|y) - \sum_y Q_Y(y)K(c|y) \right) \\
    &= \sum_y (P_Y(y) - Q_Y(y)) \sum_{c \in A} K(c|y) \\
    &= \sum_y (P_Y(y) - Q_Y(y)) K(A|y),
\end{align}

where $K(A|y) = \sum_{c \in A} K(c|y)$ is the probability of the output being in $A$ given that the input was $y$. Note that $0 \le K(A|y) \le 1$.

Let $h(y) = K(A|y)$. Then the expression becomes $\sum_y (P_Y(y) - Q_Y(y)) h(y)$.
The total variation distance can also be expressed as:
\begin{equation}
    \mathrm{TV}(P,Q) = \frac{1}{2} \sup_{h: \mathcal{Y} \to [0,1]} \left| \sum_y (P_Y(y) - Q_Y(y)) h(y) \right|.
\end{equation}
Since $h(y) = K(A|y)$ is a function from $\mathcal{Y}$ to $[0,1]$, we have:
\begin{align}
    |P_C(A) - Q_C(A)| &\le \sup_{h: \mathcal{Y} \to [0,1]} \left| \sum_y (P_Y(y) - Q_Y(y)) h(y) \right| \notag \\
    &= 2 \cdot \mathrm{TV}(P_Y, Q_Y).
\end{align}
Taking the supremum over all $A \subseteq \mathcal{C}$ on the left side gives:
\begin{equation}
    \mathrm{TV}(P_C, Q_C) = \sup_{A \subseteq \mathcal{C}} |P_C(A) - Q_C(A)| \le \mathrm{TV}(P_Y, Q_Y).
\end{equation}
This shows that processing data through a channel cannot increase the total variation distance between the distributions.
\end{proof}

\paragraph{Theorem 1 (Risk--divergence bound).}
(The statement of the theorem is in the main text)
\begin{proof}
The proof connects the risk gap to the channel divergences through a series of inequalities.

First, we express the risk gap by conditioning on the context $(X,K)$:
{\small
\begin{align}
    |L_{\mathrm{tgt}}(\theta) - L_{\mathrm{train}}(\theta)| &= \left| \mathbb{E}_{\mathrm{tgt}}[L(C;\theta)] - \mathbb{E}_{\mathrm{obs}}[L(C;\theta)] \right| \\
    &= \left| \mathbb{E}_{X,K} \left[ \mathbb{E}_{\mathrm{tgt}}[L(C;\theta)|X,K] - \mathbb{E}_{\mathrm{obs}}[L(C;\theta)|X,K] \right] \right| \\
    &\le \mathbb{E}_{X,K} \left[ \left| \mathbb{E}_{\mathrm{tgt}}[L(C;\theta)|X,K] - \mathbb{E}_{\mathrm{obs}}[L(C;\theta)|X,K] \right| \right].
\end{align}
}
The last step uses Jensen's inequality for the convex function $| \cdot |$.

Now, for a fixed $(X,K)=(x,k)$, we apply the Risk-TV inequality (Lemma 1) with $f(c) = L(c;\theta)$ and $B=L_{\max}$:
\begin{align}
    &\left| \mathbb{E}_{\mathrm{tgt}}[L(C;\theta)|x,k] - \mathbb{E}_{\mathrm{obs}}[L(C;\theta)|x,k] \right| \notag \\
    &\qquad\le 2L_{\max} \cdot \mathrm{TV}(p_{\mathrm{tgt}}(C|x,k), p_{\mathrm{obs}}(C|x,k)).
\end{align}
By assumption, the click $C$ depends on $(X,K)$ only through the bias channels $Y$. This means that the distributions $p_{\mathrm{tgt}}(C|x,k)$ and $p_{\mathrm{obs}}(C|x,k)$ are obtained by passing $p_{\mathrm{tgt}}(Y|x,k)$ and $p_{\mathrm{obs}}(Y|x,k)$ through the same Markov kernel $p(c|x,k,y)$. By the TV contraction property (Lemma 2), we have:
\begin{equation}
    \mathrm{TV}(p_{\mathrm{tgt}}(C|x,k), p_{\mathrm{obs}}(C|x,k)) \le \mathrm{TV}(p_{\mathrm{tgt}}(Y|x,k), p_{\mathrm{obs}}(Y|x,k)).
\end{equation}
Combining these inequalities and taking the expectation over $(X,K)$ gives:
{\small
\begin{equation}
    |L_{\mathrm{tgt}}(\theta) - L_{\mathrm{train}}(\theta)| \le 2L_{\max} \mathbb{E}_{X,K} \left[ \mathrm{TV}(p_{\mathrm{tgt}}(Y|X,K), p_{\mathrm{obs}}(Y|X,K)) \right].
\end{equation}
}
Now, we apply Pinsker's inequality, which states that $\mathrm{TV}(P,Q) \le \sqrt{\frac{1}{2} D_{\mathrm{KL}}(P||Q)}$:
{\small
\begin{equation}
    |L_{\mathrm{tgt}}(\theta) - L_{\mathrm{train}}(\theta)| \le 2L_{\max} \mathbb{E}_{X,K} \left[ \sqrt{\frac{1}{2} D_{\mathrm{KL}}(p_{\mathrm{obs}}(Y|X,K) || p_{\mathrm{tgt}}(Y|X,K))} \right].
\end{equation}
}
Finally, we use Jensen's inequality for the concave function $\sqrt{\cdot}$: $\mathbb{E}[\sqrt{Z}] \le \sqrt{\mathbb{E}[Z]}$. This gives the first part of the theorem:
{\small
\begin{equation}
    |L_{\mathrm{tgt}}(\theta) - L_{\mathrm{train}}(\theta)| \le 2L_{\max} \sqrt{\frac{1}{2} \mathbb{E}_{X,K} \left[ D_{\mathrm{KL}}(p_{\mathrm{obs}}(Y|X,K) || p_{\mathrm{tgt}}(Y|X,K)) \right]}.
\end{equation}
}
The second part of the theorem follows from applying the chain rule for KL divergence to the term inside the square root, which decomposes the divergence over the vector $Y$ into a sum of divergences over its components.
\end{proof}

\paragraph{Theorem 2 (Divergence--leakage bound).}
(The statement of the theorem is in the main text)
\begin{proof}
This proof establishes a connection between the channel divergence and the mutual information leakage. The key is the Strong Data-Processing Inequality (SDPI) from Assumption 1.

Fix a channel $Y \in \mathcal{Y}$ and a context $(X,K)=(x,k)$. We consider two joint distributions over $(Z,X,K,Y)$. The first, $p_{\mathrm{obs}}$, is the observational distribution from the logging data. The second, $p_{\mathrm{tgt}}$, is the target distribution. The channel from the bias sources $Z$ to the channel $Y$ is given by the Markov kernel $K_Y(y|z,x,k)$. We can write the joint distributions as:
{\small
\begin{align}
    p_{\mathrm{obs}}(z,x,k,y) &= p_{\mathrm{obs}}(z,x,k) K_Y(y|z,x,k) \\
    p_{\mathrm{tgt}}(z,x,k,y) &= p_{\mathrm{tgt}}(z,x,k) K_Y(y|z,x,k)
\end{align}
}
We now apply the SDPI (Assumption 1) conditionally on $(X,K)=(x,k)$. Let $P_{Z,X,K} = p_{\mathrm{obs}}$ and $Q_{Z,X,K} = p_{\mathrm{tgt}}$. The SDPI states that:
\begin{equation}
    D_{\mathrm{KL}}(p_{\mathrm{obs}}(Y|x,k) || p_{\mathrm{tgt}}(Y|x,k)) \le c_Y \cdot I_{\mathrm{obs}}(Z;Y | X=x, K=k).
\end{equation}
The left-hand side is exactly the definition of the channel divergence $CI_Y(x,k)$. The term $I_{\mathrm{obs}}(Z;Y | X=x, K=k)$ is the conditional mutual information under the observational distribution, which we refer to as the "leakage".

Taking the expectation of both sides with respect to $(X,K) \sim p_{\mathrm{obs}}(x,k)$ gives the final result:
\begin{equation}
    \mathbb{E}_{X,K}[CI_Y(X,K)] \le c_Y \mathbb{E}_{X,K}[I(Z;Y|X,K)].
\end{equation}
This shows that the expected channel divergence is bounded by the expected leakage, scaled by the constant $c_Y$ from the SDPI.
\end{proof}

\end{document}